\pdfoutput=1
\documentclass[acmsmall,screen,nonacm]{acmart}
\setcopyright{none}
\renewcommand\footnotetextcopyrightpermission[1]{}

\usepackage{booktabs}
\usepackage{amsmath}
\usepackage{graphicx}
\usepackage{xspace}
\usepackage{multirow}
\usepackage{array}
\usepackage{xcolor}
\usepackage{tikz}
\usetikzlibrary{arrows.meta,positioning,fit,calc}
\usepackage[capitalise,nameinlink]{cleveref}

\newcommand{\baro}{BARO\xspace}
\newcommand{\circa}{CIRCA\xspace}
\newcommand{\ediag}{$\epsilon$-Diagnosis\xspace}
\newcommand{\rcd}{RCD\xspace}
\newcommand{\causalrca}{CausalRCA\xspace}
\newcommand{\cfa}{counterfactual attribution\xspace}
\newcommand{\simplerca}{SimpleRCA\xspace}
\newcommand{\tracerca}{TraceRCA\xspace}
\newcommand{\microrank}{MicroRank\xspace}
\newcommand{\maxz}{max-$|Z|$\xspace}
\newcommand{\alertcount}{alert-count\xspace}
\newcommand{\cdmin}{CD-1min\xspace}
\newcommand{\rcaeval}{RCAEval\xspace}
\newcommand{\openrca}{OpenRCA\xspace}
\newcommand{\petshop}{PetShop\xspace}
\newcommand{\acc}{Acc@1\xspace}

\newcommand{\NlongRows}{19,893\xspace}

\newcommand{\NcwDiff}{1\xspace}
\newcommand{\NReOnePubPairs}{10\xspace}
\newcommand{\NReOnePubRev}{1\xspace}

\newcommand{\NReOneAllPairs}{28\xspace}
\newcommand{\NReOneAllRev}{8\xspace}
\newcommand{\NReOneAllRevCI}{4\xspace}
\newcommand{\NOpenrcaPubPairs}{3\xspace}

\newcommand{\NOpenrcaAllPairs}{15\xspace}
\newcommand{\NOpenrcaAllRev}{7\xspace}
\newcommand{\NOpenrcaAllRevCI}{0\xspace}

\newcommand{\NPetshopPubPairs}{10\xspace}
\newcommand{\NPetshopPubRev}{5\xspace}
\newcommand{\NPetshopPubRevCI}{2\xspace}

\newcommand{\NPetshopAllPairs}{55\xspace}
\newcommand{\NPetshopAllRev}{12\xspace}
\newcommand{\NPetshopAllRevCI}{4\xspace}

\newcommand{\NReTwoPubPairs}{28\xspace}
\newcommand{\NReTwoPubRev}{10\xspace}
\newcommand{\NReTwoPubRevCI}{6\xspace}

\newcommand{\NMainPubPairs}{23\xspace}
\newcommand{\NMainPubRev}{7\xspace}
\newcommand{\NMainPubRevCI}{2\xspace}
\newcommand{\NMainAllPairs}{98\xspace}
\newcommand{\NMainAllRev}{27\xspace}
\newcommand{\NMainAllRevCI}{8\xspace}
\newcommand{\ReOneCrRcdPooled}{$-$0.030\xspace}
\newcommand{\ReOneCrRcdPooledCI}{[$-$0.077, 0.017]\xspace}
\newcommand{\ReOneCrRcdOB}{$+$0.075\xspace}
\newcommand{\ReOneCrRcdSS}{$-$0.213\xspace}
\newcommand{\ReOneCrRcdTT}{$+$0.048\xspace}
\newcommand{\ReOneCrRcdSeedZero}{$+$0.027\xspace}

\newcommand{\ReOneCrRcdSeedOne}{$-$0.035\xspace}

\newcommand{\ReOneCrRcdSeedTwo}{$-$0.083\xspace}

\newcommand{\ReOneBaroMinMargin}{0.147\xspace}
\newcommand{\ReOneBaroMaxMargin}{0.656\xspace}

\newcommand{\ExAboveRevHigh}{$-$0.038\xspace}
\newcommand{\OpenrcaBaroMean}{0.127\xspace}
\newcommand{\OpenrcaRcdMean}{0.100\xspace}
\newcommand{\OpenrcaEdiagMean}{0.041\xspace}
\newcommand{\OpenrcaBaroRcdPooled}{$+$0.027\xspace}
\newcommand{\OpenrcaBaroRcdPooledCI}{[0.002, 0.051]\xspace}
\newcommand{\OpenrcaBaroRcdTel}{$-$0.010\xspace}
\newcommand{\OpenrcaBaroRcdTelCI}{[$-$0.058, 0.035]\xspace}
\newcommand{\OpenrcaBaroRcdSeedsWithRev}{2\xspace}
\newcommand{\PetshopRcdHighHits}{0\xspace}
\newcommand{\PetshopHighN}{26\xspace}
\newcommand{\PetshopBaroRcdHigh}{$+$0.154\xspace}
\newcommand{\PetshopBaroRcdHighCI}{[0.038, 0.308]\xspace}

\newcommand{\PetshopCfaRcdHigh}{$+$0.308\xspace}
\newcommand{\PetshopCfaRcdHighCI}{[0.154, 0.500]\xspace}

\newcommand{\PetshopPubRevNmin}{3\xspace}
\newcommand{\PetshopPubRevCINmin}{1\xspace}

\newcommand{\PetshopRankedCorrMean}{0.639\xspace}

\newcommand{\NPubBelow}{8\xspace}
\newcommand{\NPubBelowRev}{6\xspace}

\newcommand{\NPubAbove}{15\xspace}

\newcommand{\NAllBelow}{34\xspace}
\newcommand{\NAllBelowRev}{25\xspace}

\newcommand{\NAllAbove}{64\xspace}
\newcommand{\NAllAboveRev}{2\xspace}

\newcommand{\NTabBelow}{31\xspace}
\newcommand{\NTabBelowRev}{27\xspace}
\newcommand{\NTabAbove}{39\xspace}
\newcommand{\NTabAboveRev}{2\xspace}
\newcommand{\NCrossPub}{18\xspace}
\newcommand{\NCrossPubAgree}{12\xspace}
\newcommand{\NCrossPubCIopp}{2\xspace}

\newcommand{\CrossBaroCircaSSOne}{$+$0.144\xspace}
\newcommand{\CrossBaroCircaSSOneCI}{[0.064, 0.232]\xspace}
\newcommand{\CrossBaroCircaSSTwo}{$-$0.711\xspace}
\newcommand{\CrossBaroCircaSSTwoCI}{[$-$0.800, $-$0.611]\xspace}

\newcommand{\CrossBaroCrSSOne}{$+$0.672\xspace}

\newcommand{\CrossBaroCrSSTwo}{$-$0.081\xspace}
\newcommand{\CrossBaroCrSSTwoCI}{[$-$0.163, $-$0.004]\xspace}

\newcommand{\NCrossPubAgreeReOneFaults}{12\xspace}
\newcommand{\RegReTwoWinnerPooled}{0.663\xspace}

\newcommand{\RegReTwoTTbest}{0.800\xspace}
\newcommand{\RegReTwoTTwinner}{0.544\xspace}

\newcommand{\RegPetshopHigh}{0.038\xspace}

\newcommand{\InSSBaroSimple}{0.848\xspace}
\newcommand{\InSSBaroRaw}{0.200\xspace}
\newcommand{\InSSCircaSimple}{0.704\xspace}
\newcommand{\InSSCircaRaw}{0.200\xspace}

\newcommand{\InTTBaroSimple}{0.560\xspace}
\newcommand{\InTTBaroRaw}{0.120\xspace}
\newcommand{\InTTCircaSimple}{0.328\xspace}
\newcommand{\InTTCircaRaw}{0.000\xspace}

\newcommand{\InReOnePubComp}{12\xspace}
\newcommand{\InReOnePubSign}{5\xspace}

\newcommand{\InReOnePubRevRaw}{2\xspace}
\newcommand{\InReOnePubRevCIRaw}{1\xspace}

\newcommand{\InReOneAllComp}{42\xspace}
\newcommand{\InReOneAllSign}{16\xspace}

\newcommand{\InReTwoComp}{18\xspace}
\newcommand{\InReTwoSign}{5\xspace}
\newcommand{\InCircaRcdSSOk}{$+$0.315\xspace}
\newcommand{\InCircaRcdSSOkCI}{[0.195, 0.432]\xspace}
\newcommand{\InCircaRcdSSRaw}{$-$0.096\xspace}
\newcommand{\InCircaRcdSSRawCI}{[$-$0.184, $-$0.005]\xspace}
\newcommand{\InCircaRcdTTOk}{$+$0.240\xspace}
\newcommand{\InCircaRcdTTOkCI}{[0.149, 0.328]\xspace}
\newcommand{\InCircaRcdTTRaw}{$-$0.091\xspace}
\newcommand{\InCircaRcdTTRawCI}{[$-$0.139, $-$0.048]\xspace}

\newcommand{\DefPubComp}{30\xspace}
\newcommand{\DefPubSign}{2\xspace}
\newcommand{\DefPubRevShipOnly}{2\xspace}
\newcommand{\DefPubRevCIShipOnly}{1\xspace}

\newcommand{\DefCircaFixed}{0.672\xspace}
\newcommand{\DefCircaShipped}{0.432\xspace}
\newcommand{\DefCircaFixedAff}{0.820\xspace}
\newcommand{\DefCircaShippedAff}{0.220\xspace}
\newcommand{\DefCrCircaFixed}{$-$0.056\xspace}
\newcommand{\DefCrCircaFixedCI}{[$-$0.144, 0.035]\xspace}
\newcommand{\DefCrCircaShipped}{$+$0.184\xspace}
\newcommand{\DefCrCircaShippedCI}{[0.077, 0.291]\xspace}
\newcommand{\DefCircaRcdFixed}{$+$0.131\xspace}
\newcommand{\DefCircaRcdShipped}{$-$0.115\xspace}
\newcommand{\CovTraceValid}{73\xspace}
\newcommand{\CovTraceAll}{0.533\xspace}

\newcommand{\CovCrReTwoTTSeedZero}{0.267\xspace}
\newcommand{\CovCrReTwoTTSeedOne}{0.133\xspace}
\newcommand{\CovCrReTwoTTSeedTwo}{0.222\xspace}

\newcommand{\CovCrSeedOneValid}{47\xspace}
\newcommand{\CovEdTelValid}{19\xspace}
\newcommand{\CovEdTelN}{51\xspace}

\newcommand{\NCovBelowOne}{25\xspace}
\newcommand{\NCovCells}{149\xspace}
\newcommand{\DenReTwoPubComp}{71\xspace}
\newcommand{\DenReTwoPubSign}{1\xspace}
\newcommand{\DenReTwoAllComp}{146\xspace}
\newcommand{\DenReTwoAllSign}{2\xspace}

\newcommand{\DenOtherComp}{364\xspace}
\newcommand{\DenCircaTraceAll}{$+$0.011\xspace}
\newcommand{\DenCircaTraceValid}{$-$0.113\xspace}

\newcommand{\FwOpenrcaPubComp}{12\xspace}
\newcommand{\FwOpenrcaPubSign}{3\xspace}

\newcommand{\FwOpenrcaAllComp}{60\xspace}
\newcommand{\FwOpenrcaAllSign}{13\xspace}

\newcommand{\FwPetshopPubComp}{40\xspace}
\newcommand{\FwPetshopPubSign}{1\xspace}

\newcommand{\FwPetshopAllComp}{220\xspace}
\newcommand{\FwPetshopAllSign}{2\xspace}

\newcommand{\ScStrictPubComp}{12\xspace}
\newcommand{\ScStrictPubSign}{4\xspace}

\newcommand{\ScStrictAllComp}{60\xspace}
\newcommand{\ScStrictAllSign}{11\xspace}

\newcommand{\ScHalfPubSign}{0\xspace}

\newcommand{\ScHalfAllSign}{3\xspace}

\newcommand{\ScStrictPubToZero}{4\xspace}
\newcommand{\ScMktOneBaroStrict}{0.000\xspace}
\newcommand{\ScMktOneBaroPartial}{0.057\xspace}
\newcommand{\ScMktOneRcdStrict}{0.000\xspace}
\newcommand{\ScMktOneRcdPartial}{0.051\xspace}
\newcommand{\ScMktOneEdiagStrict}{0.000\xspace}
\newcommand{\ScMktOneEdiagPartial}{0.038\xspace}
\newcommand{\LinePooled}{$+$0.135\xspace}
\newcommand{\LinePooledCI}{[$-$0.041, 0.300]\xspace}

\newcommand{\LineWins}{3\xspace}
\newcommand{\LineLosses}{1\xspace}
\newcommand{\LineTies}{0\xspace}
\newcommand{\LineCovBaro}{1.00\xspace}
\newcommand{\LineCovRcd}{1.00\xspace}
\newcommand{\LineWorst}{$-$0.154\xspace}
\newcommand{\LineWorstCI}{[$-$0.308, $-$0.038]\xspace}
\newcommand{\NBonfComp}{82\xspace}
\newcommand{\NBonfRevCI}{1\xspace}
\newcommand{\NBonfCIExcl}{46\xspace}
\newcommand{\NBonfCIAdjExcl}{31\xspace}
\newcommand{\SurveyN}{23\xspace}
\newcommand{\SurveyMulti}{17\xspace}

\newcommand{\SurveyRelAny}{7\xspace}
\newcommand{\SurveyRelDemo}{3\xspace}
\newcommand{\SurveyRelOwn}{3\xspace}
\newcommand{\SurveyRelUndet}{2\xspace}
\newcommand{\SurveyRelFullCheckable}{21\xspace}

\newcommand{\XNMethods}{6\xspace}
\newcommand{\XNPairs}{15\xspace}

\newcommand{\XNBothSigns}{10\xspace}
\newcommand{\XNOneSign}{5\xspace}

\newcommand{\XLofoN}{30\xspace}
\newcommand{\XLofoBoth}{28\xspace}

\newcommand{\XLksoN}{550\xspace}
\newcommand{\XLksoBoth}{549\xspace}

\newcommand{\HNDec}{36\xspace}
\newcommand{\HErr}{8\xspace}
\newcommand{\HTies}{2\xspace}
\newcommand{\HStrict}{2\xspace}

\newcommand{\HRec}{19\xspace}
\newcommand{\HRecErr}{4\xspace}
\newcommand{\HFlag}{17\xspace}
\newcommand{\HFlagErr}{4\xspace}

\newcommand{\LosoRegPubN}{3\xspace}
\newcommand{\LosoRegPubOf}{11\xspace}
\newcommand{\LosoRegPubMax}{0.125\xspace}
\newcommand{\LosoRegAllN}{9\xspace}
\newcommand{\LosoRegAllOf}{11\xspace}
\newcommand{\LosoRegAllMax}{0.375\xspace}
\newcommand{\NAudPubPairs}{20\xspace}
\newcommand{\NAudPubRev}{6\xspace}
\newcommand{\NAudPubRevCI}{2\xspace}

\newcommand{\NMainPubEdiagPairs}{10\xspace}
\newcommand{\NMainPubRevNmin}{5\xspace}
\newcommand{\NMainPubRevCINmin}{1\xspace}

\newcommand{\ExAboveRevHighCI}{[$-$0.269, 0.192]\xspace}
\newcommand{\NPubAboveHold}{14\xspace}

\newcommand{\HDecReOne}{12\xspace}
\newcommand{\HErrReOne}{3\xspace}
\newcommand{\HDecOpenrca}{8\xspace}

\newcommand{\HDecPetshop}{16\xspace}

\newcommand{\HErrPetshopHigh}{2\xspace}
\newcommand{\HErrPetshopTmpOne}{2\xspace}

\newcommand{\HEdiagDec}{11\xspace}
\newcommand{\NCrossPubChange}{6\xspace}
\newcommand{\NCrossPubChangeBaro}{5\xspace}
\newcommand{\CrossCrEdSSOne}{$+$0.032\xspace}
\newcommand{\CrossCrEdSSOneCI}{[$-$0.051, 0.117]\xspace}
\newcommand{\CrossCrEdSSTwo}{$-$0.007\xspace}
\newcommand{\CrossCrEdSSTwoCI}{[$-$0.104, 0.089]\xspace}
\newcommand{\ScOpenrcaOfficialMin}{0.036\xspace}
\newcommand{\ScOpenrcaOfficialMax}{0.242\xspace}
\newcommand{\ScOpenrcaStrictMin}{0.000\xspace}
\newcommand{\ScOpenrcaStrictMax}{0.157\xspace}
\newcommand{\ScStrictAllRevNonzero}{7\xspace}
\newcommand{\ScStrictAllRevToZero}{1\xspace}
\newcommand{\InReOnePubRevRawOB}{1\xspace}
\newcommand{\SeedReOnePubComp}{21\xspace}
\newcommand{\SeedReOnePubSign}{3\xspace}

\newcommand{\SeedOpenrcaPubComp}{8\xspace}
\newcommand{\SeedOpenrcaPubSign}{3\xspace}

\newcommand{\SeedReTwoPubComp}{19\xspace}
\newcommand{\SeedReTwoPubSign}{2\xspace}

\newcommand{\BaroReTwoOBTopMetricN}{77\xspace}
\newcommand{\BaroReTwoOBAccThree}{0.878\xspace}
\newcommand{\BaroReTwoSSTopMetricN}{81\xspace}
\newcommand{\BaroReTwoSSAccThree}{0.956\xspace}
\newcommand{\BaroAccReOneOB}{0.736\xspace}
\newcommand{\BaroAccReOneSS}{0.848\xspace}
\newcommand{\BaroAccReTwoOB}{0.144\xspace}
\newcommand{\BaroAccReTwoSS}{0.067\xspace}
\newcommand{\ReOneOBColsMin}{51\xspace}
\newcommand{\ReOneOBColsMax}{60\xspace}
\newcommand{\ReOneSSColsMin}{58\xspace}
\newcommand{\ReOneSSColsMax}{64\xspace}
\newcommand{\ReTwoOBColsMin}{70\xspace}
\newcommand{\ReTwoOBColsMax}{78\xspace}

\newcommand{\ReOneSSRawColsMin}{421\xspace}
\newcommand{\ReOneSSRawColsMax}{439\xspace}
\newcommand{\ReOneTTRawColsMin}{1,180\xspace}
\newcommand{\ReOneTTRawColsMax}{1,446\xspace}
\newcommand{\ReOneTTColsMin}{199\xspace}
\newcommand{\ReOneTTColsMax}{239\xspace}
\newcommand{\ReOneSSRows}{721\xspace}
\newcommand{\ReTwoRows}{1,441\xspace}

\begin{document}

\title{What a Pooled Leaderboard Does Not Say: Auditing the Scope and Provenance of Method Comparisons in Microservice RCA Benchmarks}

\author{Lining Hu}
\email{ring.hu@sjtu.edu.cn}
\affiliation{\department{School of Electronic Information and Electrical Engineering}\institution{Shanghai Jiao Tong University}\city{Shanghai}\country{China}}
\author{Ting Liu}
\email{louisa\_liu@sjtu.edu.cn}
\affiliation{\department{School of Electronic Information and Electrical Engineering}\institution{Shanghai Jiao Tong University}\city{Shanghai}\country{China}}
\author{Yuzhuo Fu}
\authornote{Corresponding author.}
\email{yzfu@sjtu.edu.cn}
\affiliation{\department{School of Electronic Information and Electrical Engineering}\institution{Shanghai Jiao Tong University}\city{Shanghai}\country{China}}

\begin{abstract}
Offline benchmarks for microservice root cause analysis (RCA) rank methods by a score pooled over several applications, and a pooled lead leaves two things unstated: on which of the systems it holds, and how the score was produced.
We audit both on three public benchmark families (\openrca, \rcaeval, and \petshop), running the released implementations of 16 methods under recorded configurations on 11 subsystems and on a second release of three of the applications.
Whether a pooled lead between published methods holds on every subsystem is decided by a quantity the leaderboard withholds, the spread of the lead across subsystems: of 15 leads whose margin exceeds this spread, 14 hold everywhere; of the 8 whose margin is below it, 6 reverse.
The per-fault and per-scenario tables that the benchmark papers publish split the same way, and every error of a ranking computed without one subsystem falls on a lead of the second kind.
On the second release a third of the comparisons change sign, most through one method's response to the metric kinds it records, and on several subsystems a simple probe or a shipped baseline beats every published method.
How the score was produced matters as much: the input table that one released runner reads is not the one its documentation describes, and this choice alone reverses a comparison between published methods with intervals excluding zero both ways; an unpinned dependency and a duplicated column header each make a method return its input column order without an error.
Of 23 RCA papers we examined, none releases the per-case outputs needed to check either premise.
We recommend that a pooled comparison state the subsystems it covers and its spread across them, the input table, the handling of failed outputs, and the denominator, and be accompanied by per-case outputs, as ours are in our replication package.
\end{abstract}

\maketitle

\section{Introduction}
\label{sec:intro}

\begin{figure}[t]
\centering
\includegraphics[width=\textwidth]{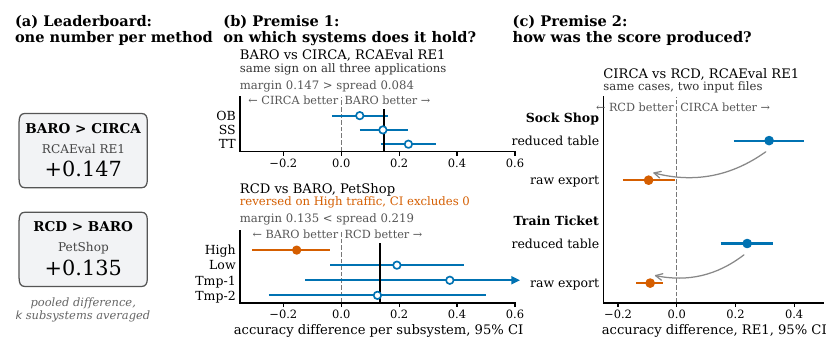}
\caption{(a) Two pooled comparisons of similar size from two benchmarks. (b) The same comparisons split by subsystem (OB, SS, TT: \rcaeval's Online Boutique, Sock Shop, and Train Ticket): points are subsystem differences $\Delta_s$ with 95\% bootstrap CIs, the vertical bar is the pooled difference, labels name the method each side favours, orange marks a subsystem whose sign is opposite to the pooled one, filled when its CI excludes 0, and each group states its pooled margin next to the between-subsystem spread of $\Delta_s$. (c) On the same \rcaeval cases, the two input tables that the release ships reverse \circa vs \rcd on two applications, with CIs excluding 0 under both.}
\label{fig:teaser}
\end{figure}

When an offline benchmark for microservice root cause analysis (RCA) reports that method $A$ outperforms method $B$, what does the claim apply to?
Benchmarks such as \openrca~\citep{xu2025openrca}, \rcaeval~\citep{pham2024rcaeval}, and \petshop~\citep{hardt2024petshop} each contain several applications or traffic scenarios, which we call subsystems, and methods are ranked by a score pooled over them.
A pooled comparison can be valid for the evaluated mixture without establishing an advantage on each constituent system; yet a method is adopted for one system, and a pooled ranking is read as a guide to that choice.
The number on the leaderboard does not say whether the advantage holds on the system at hand.
It leaves two premises unstated: on which systems the advantage holds, and how the score was produced.
\Cref{fig:teaser} shows one instance of each.
Two pooled leads of similar size (a) differ once split by subsystem (b): one has the same sign on every application, the other is reversed on one scenario with a confidence interval (CI) that excludes zero, and what separates them is the spread of the lead across subsystems, which the leaderboard does not report.
On the same cases, reading the input table that the released runner selects instead of the one its documentation describes (c) reverses a comparison between two published methods, with CIs excluding zero both ways.

Three observations motivate a systematic check.
Papers report per-system results but state superiority over the whole benchmark, and the per-case outputs needed to check such a claim with its uncertainty are rarely released.
The releases leave choices to whoever runs them, and some of these choices change scores without raising an error: \rcaeval's runner reads the one of its two metric tables that its documentation does not describe, one method returns its input column order under a newer version of a dependency the release does not pin, and a duplicated column header in 50 input tables makes another method fail in the same silent way.
Recent work asks whether these benchmarks are hard enough~\citep{fang2025simplerca}, but not which of the comparisons drawn from them hold across their subsystems.

We audit three public benchmark families: 11 subsystems in the main analysis, and a second, independently collected release of three of the applications.
We run the released implementations of 16 methods (10 published methods, 3 simple probes, and 3 baselines shipped by a benchmark) under recorded configurations and score them case by case on every case of the subsystems where they run.
We compare these runs, not the numbers the original papers report; on \openrca, where no published comparison of these methods exists, the comparison is ours (\cref{sec:design-bench}).
We ask one question about each premise.
\textbf{RQ1 (scope):} how stable are comparisons between published methods across the subsystems of a benchmark, and across a second release of the same applications?
\textbf{RQ2 (provenance):} which choices in the released pipeline---the input representation, the handling of failed outputs, the denominator, and the scoring and weighting rule---change these comparisons, and by how much?

Our findings are:
\begin{itemize}
\item \textbf{Within a benchmark, a pooled lead between published methods holds on every subsystem when its margin exceeds the spread of the lead across subsystems, and mostly reverses when it does not; the leaderboard reports the margin and withholds the spread.} Of \NPubAbove such leads with margin above spread, \NPubAboveHold hold on every subsystem; of the \NPubBelow below it, \NPubBelowRev reverse, among them both reversals with CI support. With probes and shipped baselines the counts are \NAllBelowRev of \NAllBelow against \NAllAboveRev of \NAllAbove, and in the per-fault and per-scenario tables the benchmark papers publish, \NTabBelowRev of \NTabBelow against \NTabAboveRev of \NTabAbove. A ranking computed on all subsystems but one picks the less accurate of two adjacent methods on the held-out subsystem in \HErr of \HNDec decisions, every one on a lead whose margin is below its spread. On several subsystems a probe, or a baseline shipped by the benchmark itself, is more accurate than every published method.
\item \textbf{How the score was produced can reverse it.} The largest change between the two releases of the same applications is one method's response to the metric kinds the second release records: of \NCrossPub comparisons repeated on it, \NCrossPubChange change sign, \NCrossPubChangeBaro of them involving \baro, which ranks a disk I/O metric first that the first release does not contain. On the same cases, the input table that the released runner reads reverses \circa against \rcd on two applications, with CIs excluding zero in both directions. Two failures leave no trace on the leaderboard. Under a drifted dependency one method returns its input column order on every case. A duplicated column header makes another method do the same, and produces the only CI-supported reversal between published methods on \rcaeval's first release; it disappears once the header is removed. The denominator, the fault composition, and the scoring rule move only differences that are already close to zero.
\item \textbf{Published papers do not release what is needed to check either premise.} Of \SurveyN RCA papers we examined (\cref{sec:bg-survey}), every one that evaluates more than one system reports per-system results. \SurveyRelAny release per-case outputs of at least one method, \SurveyRelDemo of them only for a single demonstration case. None of the \SurveyRelFullCheckable whose releases we could reach releases enough to recompute every paired comparison of its main results without rerunning the methods.
\end{itemize}

This paper contributes:
(i) a matched per-case table over three RCA benchmark families and 16 methods, with the configuration of every run, and two integrity checks for released pipelines that anyone can apply: whether the runner reads the input table that the documentation describes, and whether a method's ranking equals its input column order;
(ii) a quantification of the two unstated premises, the scope of pooled comparisons across subsystems, a second release, and held-out subsystems, and the provenance of scores across the recorded metric kinds and four stages of the released pipelines; and
(iii) a reporting format that states both premises, with recommendations to benchmark maintainers tied to the problems we verified.
We do not propose an RCA method.

\section{Background and Research Questions}
\label{sec:background}

\subsection{Offline RCA Benchmarks}
\label{sec:bg-bench}

In microservice root cause analysis, a method receives telemetry from a window around a failure (metrics and, in some benchmarks, logs and traces) and returns a ranked list of candidate root causes, usually services or components.
A case is scored by whether the true root cause is ranked first (\acc) or, in \openrca, by the fraction of required answer items that are correct.
We audit offline benchmarks, which release fault cases, ground truth, and a scoring rule so that every method can be scored on the same cases (\openrca~\citep{xu2025openrca}, \rcaeval~\citep{pham2024rcaeval}, \petshop~\citep{hardt2024petshop}); single-application testbeds such as Train Ticket~\citep{zhou2018trainticket}, competitions, and interactive benchmarks in which an agent queries a live system or a recorded snapshot through diagnostic tools~\citep{aiopslab,jha2025itbench,wang2026cloudopsbench} are outside its scope.

An offline benchmark is released as per-case data tables, a runner that loads them and executes the methods, and a scoring script.
A pooled score is the product of that pipeline: which table the runner reads, what happens when a method fails on a case, which cases enter the denominator, and how cases are scored and weighted.
The benchmark papers state the scoring rule; the other choices are made in code.
Each benchmark also contains several applications or traffic scenarios, which we call subsystems; its paper and the method papers that use it report a score per subsystem and a score pooled over all of them, and it is the pooled score that is ranked.

\subsection{Research Questions}
\label{sec:rqs}

A pooled leaderboard reports one number per method. Reading it as evidence that method $A$ should be preferred to method $B$ assumes two things the number does not state: on which systems the advantage holds, and how the score was produced. We ask one question about each.
\textbf{RQ1 (scope):} how stable are comparisons between published methods across the subsystems of a benchmark, and across a second, independently collected release of the same applications?
\textbf{RQ2 (provenance):} which choices in the released evaluation pipeline---the input representation, the handling of failed outputs, the denominator, and the scoring and weighting rule---change these comparisons, and by how much?

\section{Audit Design}
\label{sec:design}

\begin{figure}[t]
\centering
\includegraphics[width=\textwidth]{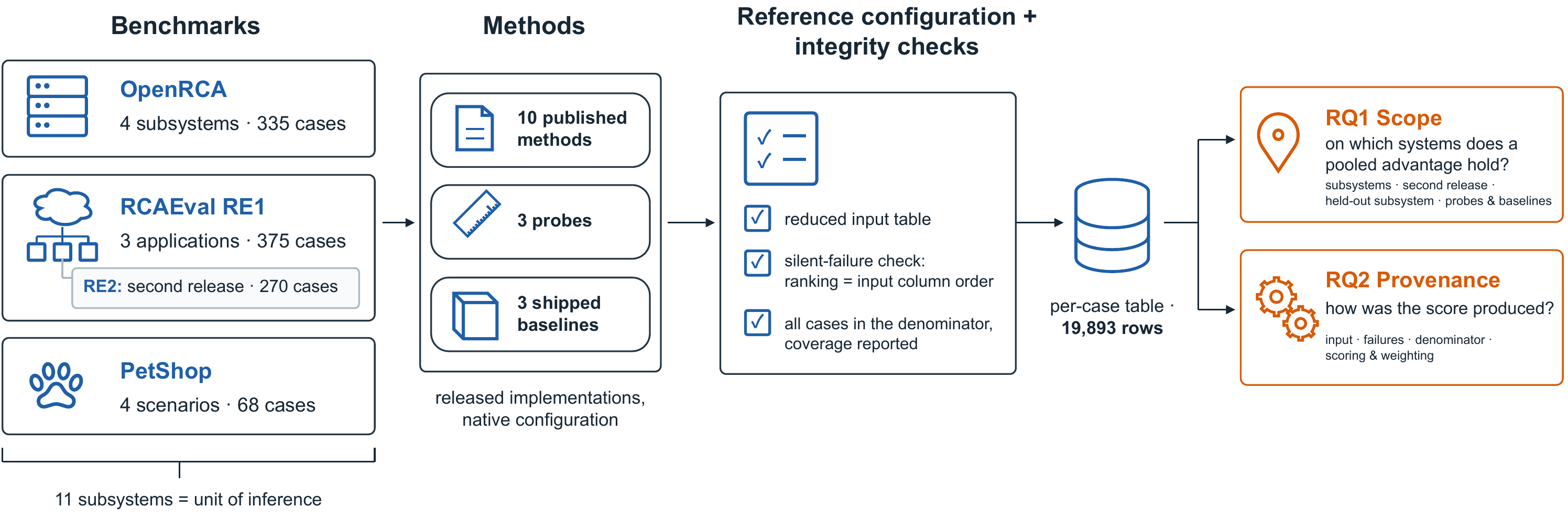}
\caption{Overview of the audit. Methods are scored case by case on every subsystem where they run; RQ1 asks on which subsystems a pooled advantage holds, RQ2 which stages of the released evaluation pipeline change it.}
\label{fig:overview}
\end{figure}

\Cref{fig:overview} summarizes the audit: three benchmark families with eleven subsystems and a second release of three of them, sixteen methods in three roles, one reference configuration with integrity checks, and the two questions asked of the resulting per-case table.

\subsection{Benchmarks and Subsystems}
\label{sec:design-bench}

We audit three public benchmark families; \cref{tab:main} lists their subsystems with the number of cases of each.
A \emph{subsystem} is one application or scenario inside a family. It is the unit at which an operator would adopt a method, and it is our unit of comparison.

\textbf{\openrca}~\citep{xu2025openrca} contains four subsystems (Bank, Market-1, Market-2, Telecom).
Each query asks for one or more of the root-cause component, the failure reason, and the occurrence time; the official scorer awards the fraction of the required items that match (component and reason by exact string, time within 60\,s).
The audited methods rank metrics, so we wrap them.
A method ranks the metrics of the query window (a 4-hour baseline plus 30 minutes on either side of the window midpoint).
The answer component is the first ranked metric that belongs to the query's candidate set, the reason is mapped from that metric's type, and the time is the moment of largest deviation.
For a query that states $n$ failures, the wrapper returns the first $n$ distinct candidate components.
No published work compares these methods on \openrca, whose own leaderboard ranks LLM agents; the comparison is our construction through this wrapper, and RQ1 marks it as such.
\cdmin runs on one-minute data with a detection window that ends one minute after the query midpoint, the setting for which it was written.

\textbf{\rcaeval}~\citep{pham2024rcaeval} releases three suites over the same three applications (Online Boutique, Sock Shop, Train Ticket).
RE1 (metrics only; five fault types) provides three subsystems of the main analysis.
RE2 (six fault types; metrics, logs, and traces) is a second, independently collected release of the same applications; we use it only to ask whether conclusions carry over between releases, and do not count it as further subsystems.
The scoring rule is top-1 at service level, a ranked metric mapping to the service named by its prefix, with the benchmark's default window of 20 minutes around the injection time.

\textbf{\petshop}~\citep{hardt2024petshop} provides four traffic scenarios (high traffic, low traffic, and two temporal scenarios), scored by its official harness at node level.

The main analysis thus has $k=11$ subsystems and 778 cases, and each method is scored on every case of the subsystems where it runs; RE2 adds 270 cases.

\subsection{Methods and Their Roles}
\label{sec:design-methods}

Methods enter the audit in one of three roles.
\emph{Published methods} are the objects of the audit: \baro~\citep{pham2024baro}, \circa~\citep{li2022circa}, \ediag~\citep{ediagnosis}, \rcd~\citep{ikram2022rcd}, and \causalrca~\citep{causalrca} on RE1; on RE2 additionally \simplerca~\citep{fang2025simplerca}, \tracerca~\citep{li2021tracerca}, \microrank~\citep{yu2021microrank}, and the multi-source variant of \baro shipped with \rcaeval; on \openrca \baro, \ediag, and \rcd; on \petshop \circa, \rcd, \ediag, \cfa~\citep{budhathoki2022causal}, and \baro.
We run the implementations that the benchmarks ship, at the release current when the audit began; the version is recorded in the replication package.
For \simplerca we use the \rcaeval branch of the authors' released code, which reproduces to two decimals every value the authors report for \simplerca and their \baro baseline on RE2 and on RE3, a third release of the same applications that we do not otherwise use.
On RE1, which provides only metrics, that branch reduces to its \baro-based metric ranking, so we do not count it as a separate method there.

\emph{Probes} are three simple rules that we add to test whether a leaderboard orders even simple alternatives consistently.
\maxz picks the metric with the largest absolute z-score after the fault relative to the pre-fault interval; \alertcount picks the service with the most metrics whose z-score exceeds 3; \cdmin, a threshold detector written for \openrca at one-minute resolution, flags metrics whose post-onset peak exceeds the baseline in both z-score and relative change and ranks components by their largest relative change.
\emph{Shipped baselines} are the traversal, ranked-correlation, and random-walk baselines that \petshop distributes.

Not every method runs on every family: \microrank and \tracerca need traces, which only RE2-OB and RE2-TT provide; \causalrca would need an estimated 4,900 GPU-hours on \openrca; \circa's causal-discovery step fails on \openrca because the metric correlation matrix is singular when there are more metrics than time points, and making it run would mean choosing a metric subset; and MicroCause, also shipped with \rcaeval, took more than 10 minutes per case and was not run.
\rcd and \causalrca are randomized; we run three seeds and average each case's score over seeds, except for the \rcd implementation that \petshop ships, which we run once through its harness.

\subsection{Reference Configuration and Integrity Checks}
\label{sec:design-protocol}

A released benchmark can be run in more than one way that is consistent with its documentation, and some of those ways change scores without any visible error.
Before comparing methods we therefore fix a reference configuration and check the outputs it produces; the alternatives set aside here become the conditions of RQ2.

\textbf{Input representation.}
RE1-SS and RE1-TT ship two metric tables per case: a raw export with several hundred to over a thousand columns, and a reduced table with a few dozen to about two hundred.\footnote{The raw export is \texttt{data.csv} (\ReOneSSRawColsMin--\ReOneSSRawColsMax columns on SS, \ReOneTTRawColsMin--\ReOneTTRawColsMax on TT) and the reduced table \texttt{simple\_data.csv} (\ReOneSSColsMin--\ReOneSSColsMax and \ReOneTTColsMin--\ReOneTTColsMax columns). The two tables of a case cover the same time points; the reduced table keeps a few aggregated metrics per service (CPU, memory, workload, latency percentiles, errors), the raw export the per-container series they are derived from. RE1-OB ships a single table of \ReOneOBColsMin--\ReOneOBColsMax columns, which we use. RE2 ships the reduced table \texttt{simple\_metrics.csv}.}
The released runner reads the raw export, and the code that would switch to the reduced table is commented out; the metric counts in the repository's documentation, however, match the reduced table.
The choice matters: \baro's Avg@5 (the metric its authors report) on the reduced tables is 0.960 (SS) and 0.779 (TT), close to the 0.95 and 0.81 that its authors, who also built \rcaeval, report, but 0.576 and 0.261 on the raw exports.
\baro is the only method whose reported values identify the file in this way, so we take the reduced tables as the reference on these two grounds; the raw exports remain an equally literal reading of the runner, and \cref{sec:rq2-input} compares every method's outputs on both.

\textbf{Silent failures.}
A method that fails inside the pipeline may still return a complete ranking, and nothing in the output marks it.
We check for the simplest signature of such a failure, a ranking equal to the input column order, and found two sources of it in the audited release.
The first is dependency drift: \causalrca calls a PageRank function that scikit-network removed in version 0.32, the released requirements set only a lower bound on that library, and the resulting exception is caught and the input column order returned.
Under scikit-network 0.33.0, \causalrca's ranking equalled the column order in 30 of 30 cases of a permutation test; under 0.31.0 in 0 of 30, and we run it under that version.
The second is a duplicated column header in 50 of the 125 RE1-OB tables, on which \circa fails and returns the column order, and \ediag ranks the duplicated column first, in 49 cases each.
Both are fixed in the current upstream release.
We apply the same fix to the input, rerun the affected methods on the affected cases, and keep the as-shipped outputs for \cref{sec:rq2-defect}.

\textbf{Conventions.}
All RE1 and RE2 methods use the 20-minute default window.
An output is invalid if the method raised an error, returned no candidate, or, on \openrca, the wrapper fell back to a fixed default answer.
Invalid outputs score 0 and the denominator is always the full case set.
We mark every accuracy whose coverage (the share of valid outputs) is below 1 and report the coverage values in \cref{sec:rq2-denominator} and the replication package.
All per-case outputs, each with the configuration of the run that produced it, are merged into one table of \NlongRows rows; after freezing, every analysis reads only this table.

\subsection{Statistics}
\label{sec:design-stats}

For methods $A$ and $B$ on subsystem $s$ with $n_s$ matched cases, the subsystem effect is the mean paired difference $\Delta_s = n_s^{-1}\sum_i (y^A_{s,i} - y^B_{s,i})$, where $y$ is the case score (top-1 hit, or the \openrca partial score).
Its 95\% confidence interval (CI) is a percentile bootstrap over cases within the subsystem ($B = 10{,}000$)~\citep{efron1993bootstrap}.
The \emph{pooled} difference $\bar\Delta$ is the unweighted mean of $\Delta_s$ over the $k$ subsystems of a family, so that each subsystem counts once, and its CI averages the subsystems' bootstrap replicates with the same weights; weighting by cases instead changes the reversal count of \NcwDiff pair only.

A \emph{point-estimate reversal} is a subsystem whose $\Delta_s$ has the opposite sign of a non-zero pooled difference; a \emph{CI-supported reversal} additionally has its CI entirely on the opposite side of zero.
The pooled difference itself need not exclude zero: a CI-supported reversal says that one subsystem has a nominal 95\% CI entirely on the side opposite to the pooled point estimate.
The CIs are nominal and not adjusted for the number of comparisons (\cref{sec:threats}).
A \emph{sign change} between two conditions includes a difference that becomes or stops being zero.
We also describe each comparison by its \emph{pooled margin} $|\bar\Delta|$ and its \emph{between-subsystem spread}, the sample standard deviation of $\Delta_s$ over the $k$ subsystems.
The replication package also gives, for every pair, a random-effects prediction interval~\citep{inthout2016plea} where $k \geq 4$ and a case-level test of the method-by-subsystem interaction; with $k \leq 4$ the interval rests on two degrees of freedom and almost always includes zero, so we draw no conclusion from it.

\emph{Held-out subsystems.}
To ask whether a pooled ranking chooses correctly on a subsystem it has not seen, we hold out each subsystem of a family in turn and rank the methods by their mean over the remaining subsystems.
For every adjacent pair in that ranking we check whether the higher-ranked method is also more accurate on the held-out subsystem.
The \emph{regret} of a pooled winner on subsystem $s$ is the accuracy of the best method on $s$ minus the winner's accuracy there.

\emph{Cross-family set.}
The methods with outputs on all 11 subsystems, \baro, \rcd, \ediag, and the three probes (\XNMethods methods, \XNPairs pairs; on \petshop, \rcd and \ediag are the implementations \petshop ships), are also compared across families.
Because \openrca's partial credit and top-1 accuracy are on different scales, we do not average them: we ask only whether a pair's subsystem differences have both signs, and whether that remains so when a family or two subsystems are dropped.

Throughout, we report two layers: comparisons between published methods, which are what a leaderboard is read for, and all pairs including probes and shipped baselines.
The former compare the released implementations as we ran them.

\subsection{Use of LLM Tools}
\label{sec:design-llm}

LLM sessions were used to locate the tables, passages, and repository contents behind the checks of \cref{sec:bg-survey}; every statement made there was verified by an author against the paper or the repository.
No LLM produced or scored a method output.

\section{RQ1 (Scope): On Which Systems Does a Pooled Advantage Hold?}
\label{sec:rq1}

\Cref{tab:main} gives every method's accuracy on every subsystem; cells with coverage below 1 are marked, and randomized methods are averaged over three seeds.
Within a family the best method changes from column to column: on RE1 it is \baro on two applications and \cdmin on the third; on \petshop it is a shipped baseline in all four scenarios, but not always the same one.
RQ1 asks how often a comparison that a pooled leaderboard would report holds on each of the subsystems behind it.

\begin{table}[t]
\caption{Accuracy (\acc; \openrca: official partial score) of every method on every subsystem, over all cases (invalid outputs score 0). $^\dagger$: coverage below 1. Bold: best in column. --: the method does not run there (\cref{sec:design-methods}). \rcd and \causalrca: mean of three seeds (\petshop's own \rcd: one run).}
\label{tab:main}
\footnotesize
\setlength{\tabcolsep}{4pt}
\centering
\begin{tabular*}{\textwidth}{@{\extracolsep{\fill}}lrrrrrr@{}}
\toprule
 & \multicolumn{3}{c}{RCAEval RE1} & \multicolumn{3}{c}{RCAEval RE2} \\
\cmidrule(lr){2-4} \cmidrule(lr){5-7}
$n$ & 125 & 125 & 125 & 90 & 90 & 90 \\
Method & OB & SS & TT & OB & SS & TT \\
\midrule
\multicolumn{7}{@{}l}{\textit{Published methods}} \\
BARO & {\bfseries 0.736} & 0.848 & {\bfseries 0.560} & 0.144 & 0.067 & 0.667 \\
CIRCA & 0.672 & 0.704 & 0.328 & 0.667 & 0.778 & 0.544 \\
$\epsilon$-Diagnosis & 0.032 & 0.144 & 0.000 & 0.033 & 0.156 & 0.000 \\
RCD & 0.541 & 0.389 & 0.088 & -- & -- & -- \\
CausalRCA & 0.616 & 0.176 & 0.136 & 0.207 & 0.148 & 0.207$^\dagger$ \\
SimpleRCA & -- & -- & -- & 0.467 & 0.244 & {\bfseries 0.800} \\
TraceRCA & -- & -- & -- & 0.100 & -- & 0.533$^\dagger$ \\
MicroRank & -- & -- & -- & 0.000 & -- & 0.122$^\dagger$ \\
BARO (multi-source) & -- & -- & -- & 0.167 & 0.067 & 0.689$^\dagger$ \\
\midrule
\multicolumn{7}{@{}l}{\textit{Probes}} \\
max-$|Z|$ & 0.504 & 0.744 & 0.424 & {\bfseries 0.778} & 0.844 & 0.411 \\
Alert-count & 0.392 & 0.376 & 0.048 & 0.211 & 0.389 & 0.211 \\
CD-1min & 0.728 & {\bfseries 0.928} & 0.280 & 0.733 & {\bfseries 0.944} & 0.789 \\
\bottomrule
\end{tabular*}
\par\medskip
\begin{tabular*}{\textwidth}{@{\extracolsep{\fill}}lrrrrrrrr@{}}
\toprule
 & \multicolumn{4}{c}{OpenRCA} & \multicolumn{4}{c}{PetShop} \\
\cmidrule(lr){2-5} \cmidrule(lr){6-9}
$n$ & 136 & 70 & 78 & 51 & 26 & 26 & 8 & 8 \\
Method & Bank & Mkt-1 & Mkt-2 & Tel & High & Low & Tmp-1 & Tmp-2 \\
\midrule
\multicolumn{9}{@{}l}{\textit{Published methods}} \\
BARO & 0.160 & 0.057 & 0.106 & 0.183 & 0.154 & 0.154 & 0.250 & 0.250 \\
CIRCA & -- & -- & -- & -- & 0.269 & 0.385 & 0.500 & 0.625 \\
$\epsilon$-Diagnosis & 0.036 & 0.038 & 0.044 & 0.046$^\dagger$ & 0.000$^\dagger$ & 0.000$^\dagger$ & 0.000$^\dagger$ & 0.125 \\
RCD & 0.089 & 0.051$^\dagger$ & 0.066$^\dagger$ & 0.193$^\dagger$ & 0.000 & 0.346 & 0.625 & 0.375 \\
Counterfactual & -- & -- & -- & -- & 0.308 & 0.154 & 0.125 & 0.250 \\
\midrule
\multicolumn{9}{@{}l}{\textit{Probes}} \\
max-$|Z|$ & 0.127 & 0.161 & {\bfseries 0.200} & 0.232 & 0.077 & 0.192 & 0.375 & 0.375 \\
Alert-count & {\bfseries 0.172} & 0.144 & 0.197 & {\bfseries 0.242} & 0.038 & 0.000 & 0.000 & 0.250 \\
CD-1min & 0.165$^\dagger$ & {\bfseries 0.161} & 0.192 & 0.163$^\dagger$ & 0.000 & 0.038 & 0.125 & 0.125 \\
\midrule
\multicolumn{9}{@{}l}{\textit{Shipped baselines}} \\
Traversal & -- & -- & -- & -- & 0.385$^\dagger$ & {\bfseries 0.577$^\dagger$} & 0.625 & {\bfseries 0.875} \\
Ranked correlation & -- & -- & -- & -- & {\bfseries 0.731$^\dagger$} & {\bfseries 0.577$^\dagger$} & {\bfseries 0.750} & 0.500 \\
Random walk & -- & -- & -- & -- & 0.000$^\dagger$ & 0.000$^\dagger$ & 0.125 & 0.000 \\
\bottomrule
\end{tabular*}

\end{table}

\subsection{Comparisons Between Published Methods}
\label{sec:rq1-pub}

\begin{figure}[t]
\centering
\includegraphics[width=0.88\textwidth]{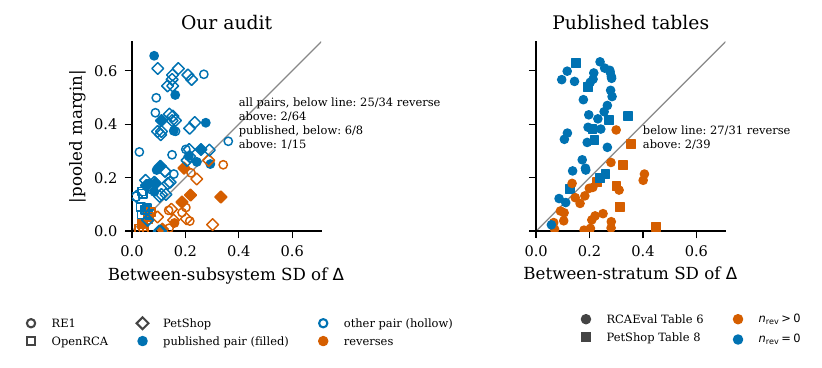}
\caption{Pooled margin $|\bar\Delta|$ against between-subsystem spread (sample standard deviation of $\Delta_s$); the diagonal is margin = spread. Left: every pair of the main analysis (filled: published methods; orange: reverses on some subsystem). Right: the pairs inside two tables the benchmark papers publish, \rcaeval's Table 6 (RE2-TT by fault type) and \petshop's Table 8 (by scenario and fault type), with the stratum in place of the subsystem.}
\label{fig:margin}
\end{figure}

\begin{figure}[t]
\centering
\includegraphics[width=\textwidth]{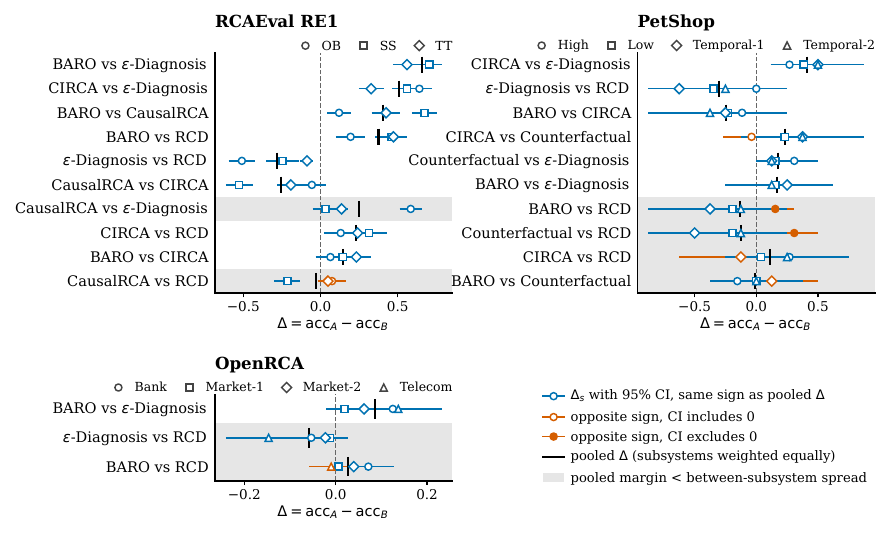}
\caption{Comparisons between published methods, by family. Each row is a pair $A$ vs $B$; markers are subsystem differences $\Delta_s$ with 95\% bootstrap CIs, the vertical tick the pooled difference (subsystems weighted equally). Orange: opposite sign to the pooled difference (filled: CI excludes 0). Grey rows: pooled margin below the between-subsystem spread, the pairs below the diagonal of \cref{fig:margin}.}
\label{fig:forest}
\end{figure}

\Cref{fig:forest} shows every comparison between two published methods that a family allows, with each subsystem's difference and its CI: \NMainPubPairs pairs, \NReOnePubPairs on RE1, \NPetshopPubPairs on \petshop, and \NOpenrcaPubPairs on \openrca, where the comparison is ours (\cref{sec:design-bench}).
\NMainPubEdiagPairs of them involve \ediag, which is never more accurate than the other method on any subsystem, and none of these reverses.
Of the \NMainPubPairs pairs, \NMainPubRev reverse on at least one subsystem and \NMainPubRevCI do so with CI support, both on one \petshop scenario (on RE1 and \petshop alone, \NAudPubRev and \NAudPubRevCI of \NAudPubPairs).
Which pairs reverse is decided by the two quantities of \cref{sec:design-stats}, the pooled margin and the between-subsystem spread of $\Delta_s$ (\cref{fig:margin}).
Of the \NPubAbove pairs whose margin exceeds the spread, \NPubAboveHold hold on every subsystem; the exception is \circa vs \cfa on \petshop, reversed on the high-traffic scenario by \ExAboveRevHigh \ExAboveRevHighCI.
Of the \NPubBelow pairs whose margin is below the spread, \NPubBelowRev reverse, among them both CI-supported reversals.
With probes and shipped baselines, \NAllBelowRev of the \NAllBelow pairs below the line reverse and \NAllAboveRev of the \NAllAbove above it.
The two tables in which the benchmark papers themselves split a score by stratum, \rcaeval's RE2-TT accuracy by fault type and \petshop's accuracy by scenario and fault type, split the same way: \NTabBelowRev of \NTabBelow pairs with margin below spread reverse on some stratum, against \NTabAboveRev of \NTabAbove above it.
A leaderboard reports the margin; the spread is what it withholds.
Dropping the two \petshop scenarios with 8 cases leaves \NMainPubRevNmin reversing pairs, \NMainPubRevCINmin with CI support, and at a Bonferroni-adjusted level \NBonfRevCI CI-supported reversal remains (\cref{sec:threats}).

\textbf{RCAEval RE1.}
Of \NReOnePubPairs pairs, \NReOnePubRev reverses, without CI support.
\baro leads each of the other four published methods on all three applications, with pooled margins from \ReOneBaroMinMargin to \ReOneBaroMaxMargin.
The one unstable pair is \causalrca vs \rcd: its pooled difference is \ReOneCrRcdPooled \ReOneCrRcdPooledCI, made of \ReOneCrRcdOB on OB, \ReOneCrRcdSS on SS, and \ReOneCrRcdTT on TT; both methods are randomized, and the pooled sign itself changes with the seed (\ReOneCrRcdSeedZero, \ReOneCrRcdSeedOne, \ReOneCrRcdSeedTwo).
This is RE1 under the reference configuration; on the files as shipped it is less stable (\cref{sec:rq2}).

\textbf{OpenRCA.}
Under the wrapper of \cref{sec:design-bench}, the three methods order as \baro (\OpenrcaBaroMean), \rcd (\OpenrcaRcdMean), \ediag (\OpenrcaEdiagMean).
\baro's lead over \rcd is \OpenrcaBaroRcdPooled \OpenrcaBaroRcdPooledCI, but Telecom reverses it (\OpenrcaBaroRcdTel \OpenrcaBaroRcdTelCI); the reversal rests on a few queries, and \OpenrcaBaroRcdSeedsWithRev of the three \rcd seeds show a reversal, on different subsystems.

\textbf{PetShop.}
The published methods are least stable here: \NPetshopPubRev of \NPetshopPubPairs pairs reverse, \NPetshopPubRevCI with CI support.
Both CI-supported reversals come from one cell: \rcd scores \PetshopRcdHighHits of \PetshopHighN on the high-traffic scenario, so \baro (\PetshopBaroRcdHigh \PetshopBaroRcdHighCI) and \cfa (\PetshopCfaRcdHigh \PetshopCfaRcdHighCI) beat it there while trailing it on pooled accuracy, with pooled CIs that include zero.
The two temporal scenarios have 8 cases each; keeping only the two scenarios with at least 25 cases leaves \PetshopPubRevNmin reversing pairs, \PetshopPubRevCINmin with CI support.

\subsection{Same Applications, Second Release}
\label{sec:rq1-release}

RE2 is a second, independently collected release of RE1's three applications, with new fault instances, a sixth fault type, and added logs and traces.
Its metric tables are also sampled differently (\ReTwoRows rows per case against \ReOneSSRows on RE1-SS) and record more metric kinds per service (\ReTwoOBColsMin--\ReTwoOBColsMax columns on OB against \ReOneOBColsMin--\ReOneOBColsMax on RE1).
The four published methods that use metrics only (\baro, \circa, \ediag, \causalrca) run with the same implementations on both releases.
Of the \NCrossPub application-level comparisons available in both releases (\cref{tab:release}), \NCrossPubAgree keep their sign and \NCrossPubChange change it.
\NCrossPubChangeBaro of the \NCrossPubChange involve \baro; the sixth, \causalrca vs \ediag on Sock Shop, moves from \CrossCrEdSSOne \CrossCrEdSSOneCI to \CrossCrEdSSTwo \CrossCrEdSSTwoCI.
\NCrossPubCIopp comparisons have CIs on opposite sides of zero, both on Sock Shop: \baro vs \circa goes from \CrossBaroCircaSSOne \CrossBaroCircaSSOneCI to \CrossBaroCircaSSTwo \CrossBaroCircaSSTwoCI, and \baro vs \causalrca from \CrossBaroCrSSOne to \CrossBaroCrSSTwo \CrossBaroCrSSTwoCI.
Restricting RE2 to RE1's five fault types leaves the count at \NCrossPubAgreeReOneFaults of \NCrossPub.
Within RE2 itself, \NReTwoPubRev of \NReTwoPubPairs published pairs reverse across the three applications (\NReTwoPubRevCI with CI support).

\baro's accuracy falls from \BaroAccReOneOB to \BaroAccReTwoOB on OB and from \BaroAccReOneSS to \BaroAccReTwoSS on SS.
Its ranking is not the input column order, every case has a valid output, and the root cause is still among its first three candidates in \BaroReTwoOBAccThree of the OB cases and \BaroReTwoSSAccThree of the SS cases; what changes is the first position.
RE2's reduced table adds disk I/O and socket metrics for every service, which RE1's tables do not contain, and \baro ranks the disk I/O of the application's database first in \BaroReTwoOBTopMetricN of the 90 OB cases and \BaroReTwoSSTopMetricN of the 90 SS cases, whatever the injected fault.
\citet{fang2025simplerca} report top-1 accuracies of 0.14 on both applications for their own \baro baseline on RE2.
What changed between the releases for this method is the set of metric kinds recorded, a part of the score's provenance that \cref{sec:rq2-summary} takes up.

\begin{table}[t]
\caption{Same application, two releases: subsystem differences $\Delta_s$ ($A$ minus $B$) between published methods on RE1 and RE2. Bold: the two releases' 95\% CIs lie on opposite sides of zero.}
\label{tab:release}
\footnotesize
\setlength{\tabcolsep}{5pt}
\begin{tabular}{@{}lrrrrrr@{}}
\toprule
 & \multicolumn{2}{c}{Online Boutique} & \multicolumn{2}{c}{Sock Shop} & \multicolumn{2}{c}{Train Ticket} \\
\cmidrule(lr){2-3} \cmidrule(lr){4-5} \cmidrule(lr){6-7}
Pair $A$ vs $B$ & RE1 & RE2 & RE1 & RE2 & RE1 & RE2 \\
\midrule
BARO vs CausalRCA & +0.120 & $-$0.063 & {\bfseries +0.672} & {\bfseries $-$0.081} & +0.424 & +0.459 \\
BARO vs CIRCA & +0.064 & $-$0.522 & {\bfseries +0.144} & {\bfseries $-$0.711} & +0.232 & +0.122 \\
BARO vs $\epsilon$-Diagnosis & +0.704 & +0.111 & +0.704 & $-$0.089 & +0.560 & +0.667 \\
CausalRCA vs CIRCA & $-$0.056 & $-$0.459 & $-$0.528 & $-$0.630 & $-$0.192 & $-$0.337 \\
CausalRCA vs $\epsilon$-Diagnosis & +0.584 & +0.174 & +0.032 & $-$0.007 & +0.136 & +0.207 \\
CIRCA vs $\epsilon$-Diagnosis & +0.640 & +0.633 & +0.560 & +0.622 & +0.328 & +0.544 \\
\bottomrule
\end{tabular}

\end{table}

\subsection{Choosing a Method for an Unseen Subsystem}
\label{sec:rq1-heldout}

A practitioner choosing a method for a new system has only the other subsystems to go by.
Holding out each of the 11 subsystems in turn and ranking the published methods of its family on the remaining ones gives \HNDec adjacent-pair decisions (RE1 \HDecReOne, \openrca \HDecOpenrca, \petshop \HDecPetshop).
In \HErr of them the higher-ranked method is less accurate on the held-out subsystem (\HStrict with a CI excluding zero), and \HTies further decisions are ties.
The \HErr wrong choices fall on pairs that \cref{sec:rq1-pub} places below the line.
On RE1 all \HErrReOne are \rcd vs \causalrca, the pair whose pooled sign changes with the seed; on \openrca the one is \baro vs \rcd on Telecom, the reversal with a CI including zero; on \petshop \HErrPetshopHigh are on the high-traffic scenario, among them \rcd vs \baro, and \HErrPetshopTmpOne on the 8-case temporal scenario.
None of the \HEdiagDec decisions that involve \ediag is wrong.
Computed on the remaining subsystems alone, margin and spread do not single out the wrong choices: \HRecErr of the \HRec decisions whose margin exceeds the spread there are wrong, and \HFlagErr of the \HFlag others (replication package).
\Cref{tab:regret} gives the cost of such a choice as regret.
The pooled winner chosen without the held-out subsystem has positive regret on \LosoRegPubN of the \LosoRegPubOf subsystems (up to \LosoRegPubMax), and with probes and shipped baselines on \LosoRegAllN of \LosoRegAllOf (up to \LosoRegAllMax).
The largest costs are concrete: on RE2, \circa has the highest pooled accuracy (\RegReTwoWinnerPooled) but scores \RegReTwoTTwinner on Train Ticket against \simplerca's \RegReTwoTTbest; on \petshop the pooled winner \circa trails another published method on the high-traffic scenario by \RegPetshopHigh.

\begin{table}[t]
\caption{Pooled winner (subsystems weighted equally) and its regret: accuracy of the best method on a subsystem minus the winner's accuracy there. All subsystems: winner chosen on all subsystems of the family. Held out: for each subsystem, winner chosen on the other subsystems only (not computed for RE2, which we use only across releases).}
\label{tab:regret}
\footnotesize
\setlength{\tabcolsep}{4pt}
\begin{tabular}{@{}lllrcrcr@{}}
\toprule
 & & & & \multicolumn{2}{c}{All subsystems} & \multicolumn{2}{c}{Held out} \\
\cmidrule(lr){5-6} \cmidrule(lr){7-8}
Family & Layer & Pooled winner & \shortstack{Winner\\pooled acc.} & \shortstack{Regret\\$> 0$} & \shortstack{Max regret\\(subsystem)} & \shortstack{Regret\\$> 0$} & Max \\
\midrule
RCAEval RE1 & Published & BARO & 0.715 & 0 of 3 & 0 & 0 of 3 & 0 \\
RCAEval RE1 & All & BARO & 0.715 & 1 of 3 & 0.080 (RE1-SS) & 2 of 3 & 0.280 \\
RCAEval RE2 & Published & CIRCA & 0.663 & 1 of 3 & 0.256 (RE2-TT) & -- & -- \\
RCAEval RE2 & All & CD-1min & 0.822 & 2 of 3 & 0.044 (RE2-OB) & -- & -- \\
OpenRCA & Published & BARO & 0.127 & 1 of 4 & 0.010 (Telecom) & 1 of 4 & 0.010 \\
OpenRCA & All & Alert-count & 0.189 & 2 of 4 & 0.017 (Market-1) & 4 of 4 & 0.078 \\
PetShop & Published & CIRCA & 0.445 & 2 of 4 & 0.125 (Temporal-1) & 2 of 4 & 0.125 \\
PetShop & All & Ranked correlation & 0.639 & 1 of 4 & 0.375 (Temporal-2) & 3 of 4 & 0.375 \\
\bottomrule
\end{tabular}

\end{table}

\subsection{Probes and Shipped Baselines}
\label{sec:rq1-all}

Adding the three probes and \petshop's shipped baselines gives \NMainAllPairs pairs in the main families, of which \NMainAllRev reverse and \NMainAllRevCI with CI support (RE1 \NReOneAllRev/\NReOneAllRevCI of \NReOneAllPairs; \openrca \NOpenrcaAllRev/\NOpenrcaAllRevCI of \NOpenrcaAllPairs; \petshop \NPetshopAllRev/\NPetshopAllRevCI of \NPetshopAllPairs).
The probes do not win everywhere or lose everywhere: \cdmin is the best method on RE1-SS but ranks below \baro on RE1-TT, and \alertcount leads \openrca's pooled ranking.
On \petshop, the shipped ranked-correlation baseline has the highest pooled accuracy (\PetshopRankedCorrMean), and its pooled advantage over every published method except \circa has a CI that excludes zero.
A leaderboard that lists only published methods therefore ranks methods that a shipped baseline of the same benchmark outperforms.

\subsection{Across Benchmark Families}
\label{sec:rq1-cross}

\XNMethods methods (\baro, \rcd, \ediag, and the three probes) have outputs on all 11 subsystems, so their comparisons can be followed across families by the signs of their subsystem differences (\cref{sec:design-stats}; per-pair tables in the replication package).
Of the \XNPairs pairs, \XNBothSigns have subsystem differences of both signs; the other \XNOneSign all involve \ediag, which is never more accurate than any of the other five on any subsystem, so a ranking that places it last holds everywhere.
Dropping any one family leaves both signs in \XLofoBoth of \XLofoN pair-by-family cases, and dropping any two subsystems in \XLksoBoth of \XLksoN: no order of the other five methods, on whatever scale the families were averaged, holds on every subsystem.

\section{RQ2 (Provenance): How the Score Was Produced}
\label{sec:rq2}

A benchmark score is the output of a pipeline, and the released pipeline leaves choices open at four stages: which representation of the telemetry the method reads, what happens when the method fails, how invalid outputs enter the denominator, and how cases are weighted and scored.
RQ1 read each subsystem difference as a property of the method and the subsystem; this section asks which choices of the released pipeline produced it.
The duplicated header of \cref{sec:design-protocol} is one such choice's effect, a subsystem anomaly that disappears once the input is fixed.
For each stage we compare the reference configuration of \cref{sec:design-protocol} with one alternative on the same cases and the same methods, and record for every pair and subsystem whether $\Delta_s$ changes sign and whether the comparison's reversal status changes (\cref{tab:factors}).
The stages are not equal: the first two produced opposite conclusions with CIs excluding zero on both sides; the last two moved, on \openrca, only differences that were already close to zero.

\begin{table}[t]
\caption{Pipeline stages that change comparison conclusions. Comp.: comparisons (one method pair on one subsystem); Sign: comparisons whose $\Delta_s$ changes sign; Rev.: comparisons that are a reversal under exactly one of the two conditions. All pairs include probes and shipped baselines; --: no published pair has outputs under both conditions. Seed: the three-seed mean of a randomized method against any single seed, over the comparisons that involve such a method (\cref{sec:rq2-denominator}).}
\label{tab:factors}
\footnotesize
\setlength{\tabcolsep}{4pt}
\begin{tabular*}{\textwidth}{@{\extracolsep{\fill}}llrrrrrr@{}}
\toprule
 & & \multicolumn{3}{c}{Published methods} & \multicolumn{3}{c}{All pairs} \\
\cmidrule(lr){3-5} \cmidrule(lr){6-8}
Factor & Family & Comp. & Sign & Rev. & Comp. & Sign & Rev. \\
\midrule
Input representation & RCAEval RE1 & 12 & 5 & 1 & 42 & 16 & 8 \\
 & RCAEval RE2 & -- & -- & -- & 18 & 5 & 7 \\
\addlinespace
Silent failure (duplicated header) & RCAEval RE1 & 30 & 2 & 2 & 84 & 3 & 4 \\
\addlinespace
Denominator (valid outputs only) & RCAEval RE1 & 30 & 0 & 0 & 84 & 0 & 0 \\
 & RCAEval RE2 & 71 & 1 & 1 & 146 & 2 & 2 \\
 & OpenRCA & 12 & 0 & 0 & 60 & 0 & 0 \\
 & PetShop & 40 & 0 & 0 & 220 & 0 & 0 \\
\addlinespace
Fault re-weighting & RCAEval RE1 & 30 & 0 & 0 & 84 & 0 & 0 \\
 & RCAEval RE2 & 71 & 0 & 0 & 146 & 0 & 0 \\
 & OpenRCA & 12 & 3 & 3 & 60 & 13 & 15 \\
 & PetShop & 40 & 1 & 1 & 220 & 2 & 1 \\
\addlinespace
Scoring (strict $=1$) & OpenRCA & 12 & 4 & 0 & 60 & 11 & 8 \\
\addlinespace
Scoring ($\geq 0.5$) & OpenRCA & 12 & 0 & 0 & 60 & 3 & 3 \\
\addlinespace
Seed (one seed vs.\ three-seed mean) & RCAEval RE1 & 21 & 3 & 4 & 39 & 5 & 6 \\
 & RCAEval RE2 & 19 & 2 & 2 & 28 & 4 & 4 \\
 & OpenRCA & 8 & 3 & 3 & 20 & 3 & 3 \\
\bottomrule
\end{tabular*}

\end{table}

\subsection{Input Representation}
\label{sec:rq2-input}

\begin{figure}[t]
\centering
\includegraphics[width=\textwidth]{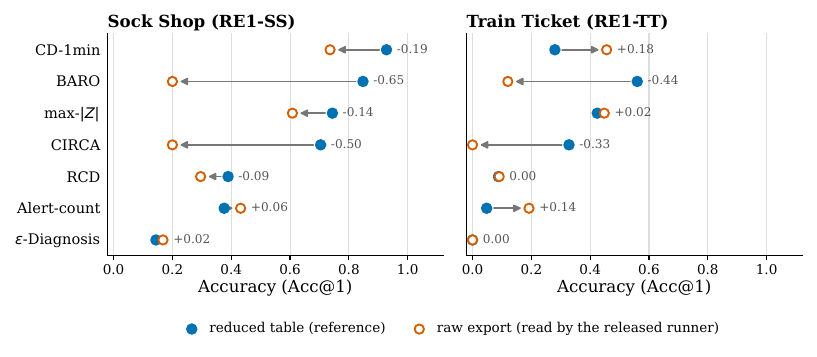}
\caption{RE1 accuracy of each method on the reduced table (the reference) and on the raw export that the released runner reads.}
\label{fig:input}
\end{figure}

RE1-SS and RE1-TT can be read from the raw export or from the reduced table (\cref{sec:design-protocol}), and the two readings do not rank the methods the same way.
\Cref{fig:input} shows each method's accuracy under both.
\baro falls from \InSSBaroSimple to \InSSBaroRaw on SS and from \InTTBaroSimple to \InTTBaroRaw on TT, and \circa from \InSSCircaSimple to \InSSCircaRaw and from \InTTCircaSimple to \InTTCircaRaw, while \alertcount and \cdmin improve on TT and \ediag and \rcd barely move.
The raw export is not a uniform penalty; it reorders the methods.
The release does not state which table its reported numbers use, and a reader who runs it as shipped obtains the raw numbers.

Among published methods, \InReOnePubSign of the \InReOnePubComp SS and TT comparisons change sign, the largest share at any stage.
\circa leads \rcd by \InCircaRcdSSOk \InCircaRcdSSOkCI on SS and \InCircaRcdTTOk \InCircaRcdTTOkCI on TT with the reduced tables; with the raw exports \rcd leads, by \InCircaRcdSSRaw \InCircaRcdSSRawCI and \InCircaRcdTTRaw \InCircaRcdTTRawCI.
All four intervals exclude zero: two analyses of the same released cases, each a literal reading of the release, report opposite conclusions with the same apparent support.
On SS the best published method changes from \baro to \rcd.
The raw exports add \InReOnePubRevRaw reversals among published pairs (\InReOnePubRevCIRaw with CI support) that are absent with the reduced tables; \InReOnePubRevRawOB of them is on OB, whose data are the same under both conditions, because the pooled sign of \circa vs \rcd changes (\cref{tab:factors} counts the SS and TT comparisons).
Among all pairs \InReOneAllSign of \InReOneAllComp comparisons change sign, each condition showing reversals the other lacks: the input representation replaces one set of conclusions with another.
RE2 ships the same two kinds of table, and there the runner reads the reduced one; the raw table changes \InReTwoSign of \InReTwoComp comparisons among the methods we could run on it.

\subsection{Silent Failures}
\label{sec:rq2-defect}

A method that fails inside the pipeline but returns a complete ranking leaves no trace on the leaderboard; the ranking is scored like any other.
Both failures found in \cref{sec:design-protocol} are of this kind.
Under the drifted dependency, \causalrca returns the input column order on every case we tested, so a comparison that involves it on the release as shipped compares the column order of the input file, not the method.
The duplicated header in 50 RE1-OB tables has an effect we can measure, because \circa and \ediag run on the same cases before and after the fix.
On the files as shipped, \circa's accuracy on RE1-OB is \DefCircaShipped; after dropping the duplicated column it is \DefCircaFixed (\DefCircaShippedAff and \DefCircaFixedAff on the 50 affected cases).
As shipped, \causalrca beats \circa on OB by \DefCrCircaShipped \DefCrCircaShippedCI, a CI-supported reversal of \circa's pooled lead; after the fix the difference is \DefCrCircaFixed \DefCrCircaFixedCI, and \circa vs \rcd on OB changes from \DefCircaRcdShipped to \DefCircaRcdFixed.
Of \DefPubComp comparisons between published methods on RE1, \DefPubSign change sign, and the shipped files produce \DefPubRevShipOnly reversals (\DefPubRevCIShipOnly with CI support) that disappear after the fix.
Neither failure raises an error or lowers coverage; the one check that reveals both is whether a method's ranking equals the order of the input columns.

\subsection{Denominator and Reported Numbers}
\label{sec:rq2-denominator}

\NCovBelowOne of \NCovCells method-by-subsystem cells (seeds counted separately) have coverage below 1, for method-specific reasons: \tracerca and \microrank fail on the same 17 RE2-TT cases with a missing-key error, leaving \CovTraceValid valid outputs of 90; \ediag returns three metrics per query, none of which belongs to the candidate set in most \openrca Telecom queries (\CovEdTelValid of \CovEdTelN valid); and one seed of \causalrca learns an empty graph in 43 RE2-TT cases (\CovCrSeedOneValid valid).
Scoring valid outputs only, instead of all cases, changes \DenReTwoPubSign of \DenReTwoPubComp comparisons between published methods on RE2 and \DenReTwoAllSign of \DenReTwoAllComp among all pairs, all on RE2-TT (\circa vs \tracerca moves from \DenCircaTraceAll to \DenCircaTraceValid), and none of the \DenOtherComp comparisons in the other families.
Two things a reported number does not state matter when it is compared with a published one: the denominator and the seed.
Our \tracerca and \microrank outputs match every per-fault value that \rcaeval reports for RE2-TT (18 cells each, within 0.01) only if the 17 failed cases are dropped, although the paper's definition of the metric divides by all cases.
Over all cases, \tracerca's \acc is \CovTraceAll instead of the reported 0.66.
\causalrca's code defines a seed but never applies it, and on RE2-TT, where \rcaeval reports 0.22, our three seeds give \CovCrReTwoTTSeedZero, \CovCrReTwoTTSeedOne, and \CovCrReTwoTTSeedTwo.
Taking any single seed instead of the three-seed mean changes the sign of \SeedReOnePubSign of the \SeedReOnePubComp comparisons between published methods that involve a randomized method on RE1, \SeedReTwoPubSign of \SeedReTwoPubComp on RE2, and \SeedOpenrcaPubSign of \SeedOpenrcaPubComp on \openrca (last row of \cref{tab:factors}).

\subsection{Fault Composition}
\label{sec:rq2-fault}

A pooled score also mixes fault types, which are not equally frequent in every subsystem.
We re-weight each subsystem so that the fault classes it shares with the rest of its family count equally.
Both \rcaeval releases contain the same number of cases for every fault type in every application, so no comparison there changes.
\openrca's four subsystems share only CPU and network faults; restricting each subsystem to these and weighting them equally changes the sign of \FwOpenrcaPubSign of \FwOpenrcaPubComp comparisons between published methods and \FwOpenrcaAllSign of \FwOpenrcaAllComp overall, none with a CI that excludes zero; on \petshop (latency and availability faults), \FwPetshopPubSign of \FwPetshopPubComp and \FwPetshopAllSign of \FwPetshopAllComp change.

\subsection{Scoring Rule}
\label{sec:rq2-scoring}

\openrca's official score gives partial credit.
Counting only fully correct answers changes the sign of \ScStrictPubSign of \ScStrictPubComp comparisons between published methods (\ScStrictAllSign of \ScStrictAllComp overall); requiring a score of at least 0.5 changes \ScHalfPubSign and \ScHalfAllSign.
All \ScStrictPubToZero sign changes between published methods under the strict rule are differences that become exactly zero, and none changes a reversal status (\cref{tab:factors}): on Market-1, \baro, \rcd, and \ediag score \ScMktOneBaroStrict, \ScMktOneRcdStrict, and \ScMktOneEdiagStrict under the strict rule and \ScMktOneBaroPartial, \ScMktOneRcdPartial, and \ScMktOneEdiagPartial with partial credit.
Under the strict rule the published methods are indistinguishable on that subsystem; the ordering reported under partial credit comes entirely from partially correct answers.
Among all pairs, \ScStrictAllRevNonzero of the reversal-status changes remain non-zero differences and \ScStrictAllRevToZero is a reversal that becomes a tie.
Every method's score on \openrca lies between \ScOpenrcaOfficialMin and \ScOpenrcaOfficialMax under the official rule and between \ScOpenrcaStrictMin and \ScOpenrcaStrictMax under the strict one, so the differences a scoring rule can move there are of this size.
Where scores are not at the floor the rule moves more: on RE2-OB, \baro's accuracy is \BaroAccReTwoOB at top-1 and \BaroReTwoOBAccThree at top-3 (\cref{sec:rq1-release}).

\subsection{Reading the Four Stages Together}
\label{sec:rq2-summary}

The first two stages changed conclusions between published methods with CI support; the last two changed, on \openrca, the sign only of differences that were close to zero in the reference analysis, or turned them into ties.
Both silent failures were found in \rcaeval, the one family whose released runner we could execute as shipped; \openrca releases no runner for these methods, and we found none in \petshop's harness.
The largest provenance effect we measured is a choice of the release rather than of the runner: which metric kinds it records (\cref{sec:rq1-release}) changed \baro's accuracy by as much as the input representation did.
The denominator, the fault weighting, and the scoring rule are choices every benchmark makes, and they are the ones a reader can usually see.
An audit should therefore start with what is least visible, which table the runner reads, which metric kinds it contains, and whether rankings equal the input column order, then ask how invalid outputs enter the denominator, and only then turn to the weighting and the scoring rule.

\section{Can Either Premise Be Checked From Published Papers?}
\label{sec:bg-survey}

The two premises can be checked from a paper only if it reports the per-system scores behind its pooled comparisons, releases enough to attach an uncertainty to them, and states how its scores were produced.
We read \SurveyN microservice RCA method and benchmark papers from 2021 to 2026, a convenience sample built from the benchmark papers, the method papers they compare, and the papers citing them for which we obtained the full text (\cref{tab:survey}), and checked these three things.
A paper counts as reporting per-system results if it gives a score for each evaluated system or dataset; a release counts as containing per-case outputs if a file in it gives a method's ranking or prediction for an identifiable case, and as complete if such files exist for every method on every dataset of the main comparison.
Scope is reported: \SurveyMulti of the \SurveyN papers evaluate more than one system, and all \SurveyMulti give a score per system, so a reader can see where a pooled lead does not hold, but not with what uncertainty.
Checking the repositories and archives linked from each paper on 2 and 3 October 2026, we found \SurveyRelAny papers whose release contains per-case outputs of at least one method.
For \SurveyRelDemo of them (\baro~\citep{pham2024baro}, \rcaeval~\citep{pham2024rcaeval}, and a study that shares \rcaeval's repository~\citep{pham2024howfar}) the outputs are those a tutorial notebook saved for a single case.
A further \SurveyRelOwn release the outputs of the authors' own method on their test cases: one archived run of \openrca's agent~\citep{xu2025openrca}, the rankings of CausalRCA~\citep{causalrca}, and the predictions of CHASE~\citep{zhao2024chase}.
One, \circa~\citep{li2022circa}, releases the rankings of every method of its simulation study, but not of its production dataset.
For \SurveyRelUndet papers a linked release had expired or could not be reached.
None of the \SurveyRelFullCheckable papers whose releases we could check releases per-case outputs for every method on every dataset of its main comparison, so for those comparisons a reader cannot compute a paired difference, its uncertainty, or a coverage figure.
\begin{table}[t]
\caption{The \SurveyN surveyed papers. Per-system: a score per evaluated system is reported (single system: one system evaluated). Per-case outputs released: what the release linked from the paper contains, checked on 2 and 3 October 2026. All methods: whether per-case outputs exist for every method on every dataset of the main comparison (partial: for every method of one experiment).}
\label{tab:survey}
\scriptsize
\setlength{\tabcolsep}{3pt}
\begin{tabular}{@{}lllllc@{}}
\toprule
Paper & Year & Venue & Per-system & Per-case outputs released & All methods \\
\midrule
BARO~\citep{pham2024baro} & 2024 & FSE 2024 & yes & one demonstration case & no \\
CIRCA~\citep{li2022circa} & 2022 & KDD 2022 & yes & all methods, simulation study only & partial \\
RCD~\citep{ikram2022rcd} & 2022 & NeurIPS 2022 & yes & -- & no \\
TraceRCA~\citep{li2021tracerca} & 2021 & IWQoS 2021 & yes & release unreachable & -- \\
RCAEval~\citep{pham2024rcaeval} & 2025 & WWW 2025 Companion & single system & one demonstration case & no \\
SimpleRCA~\citep{fang2025simplerca} & 2026 & FSE 2026 & yes & -- & no \\
OpenRCA~\citep{xu2025openrca} & 2025 & ICLR 2025 & yes & own method, all test cases & no \\
PetShop~\citep{hardt2024petshop} & 2024 & CLeaR 2024 & single system & -- & no \\
Eadro~\citep{lee2023eadro} & 2023 & ICSE 2023 & yes & -- & no \\
mABC~\citep{zhang2024mabc} & 2024 & EMNLP 2024 Findings & yes & -- & no \\
MULAN~\citep{zheng2024mulan} & 2024 & WWW 2024 & yes & -- & no \\
OCEAN~\citep{zheng2025ocean} & 2025 & IEEE BigData 2025 & yes & -- & no \\
TVDiag~\citep{xie2025tvdiag} & 2025 & TOSEM & yes & -- & no \\
CHASE~\citep{zhao2024chase} & 2024 & arXiv 2024 & yes & own method, all test cases & no \\
How Far Are We~\citep{pham2024howfar} & 2024 & ASE 2024 & yes & one demonstration case & no \\
LEMMA-RCA~\citep{zheng2024lemma} & 2024 & arXiv 2024 & yes & -- & no \\
SparseRCA~\citep{yao2024sparserca} & 2024 & ISSRE 2024 & single system & -- & no \\
TraceContrast~\citep{zhang2024tracecontrast} & 2024 & ICSE 2024 & single system & -- & no \\
TraceDiag~\citep{ding2023tracediag} & 2023 & ESEC/FSE 2023 & single system & -- & no \\
RUN~\citep{lin2024run} & 2024 & AAAI 2024 & yes & -- & no \\
DynaCausal~\citep{zhang2025dynacausal} & 2025 & arXiv 2025 & yes & -- & no \\
CausalRCA~\citep{causalrca} & 2023 & JSS 2023 & single system & own method, all test cases & no \\
Too Many Cooks~\citep{zhang2025toomanycooks} & 2025 & ISSRE 2025 & yes & release unreachable & -- \\
\bottomrule
\end{tabular}

\end{table}

Whether a number was computed on the raw or the reduced table is a question that even the benchmark's own paper does not answer.
\rcaeval's paper describes the file format of its release but not which of the two tables its reported numbers use, and \cref{sec:rq2-input} shows that the two rank the methods differently.
Superiority claims are nonetheless stated over the benchmark as a whole, for example ``Extensive experiments conducted on the synthetic and real-world microservice-based datasets demonstrate that RUN noticeably outperforms the state-of-the-art root cause analysis methods''~\citep{lin2024run}.

\section{Discussion and Recommendations}
\label{sec:discussion}

\subsection{Stating Both Premises}
\label{sec:disc-authors}

A per-subsystem table states the first premise, but pooled numbers will continue to be quoted, so we suggest that every pooled comparison carry a line of context with one part per premise:
\par\medskip\noindent\hspace*{2em}%
{\small\begin{tabular}[t]{@{}l@{\quad}l@{}}
\textit{scope} & pooled $\Delta$ [CI] \,$\cdot$\, subsystems won/tied/lost of $k$ \,$\cdot$\, worst subsystem $\Delta$ [CI] \\
\textit{provenance} & coverage \,$\cdot$\, input table \,$\cdot$\, invalid outputs
\end{tabular}}
\par\medskip\noindent
For \rcd vs \baro on \petshop this line reads: \LinePooled \LinePooledCI \,$\cdot$\, \LineWins/\LineTies/\LineLosses of 4 \,$\cdot$\, high traffic \LineWorst \LineWorstCI \,$\cdot$\, \LineCovRcd vs \LineCovBaro \,$\cdot$\, official harness \,$\cdot$\, scored 0.
A reader sees at once that the pooled CI includes zero, that the lead holds on three of four scenarios, that the one loss has a CI excluding zero, and on which input and denominator the numbers rest; the pooled number alone, \LinePooled, says none of this.
The subsystem count and the worst subsystem carry the spread that \cref{sec:rq1-pub} found to decide whether a lead holds everywhere.
The line is a reporting format, not a decision rule, and every quantity in it can be computed from per-case outputs.
Where the lead does not hold everywhere, a count such as ``leads on 3 of 4 subsystems'' says more than ``outperforms''.

\subsection{For Benchmark Maintainers}
\label{sec:disc-maintainers}

Each recommendation below corresponds to a problem we verified.

(i) The runner should name the input table and preprocessing it uses, in code and not only in documentation; the two readings of RE1 reversed a comparison between published methods with CI support on both sides (\cref{sec:rq2-input}).

(ii) Dependency versions should be pinned, and a method that fails inside the pipeline should fail visibly rather than return a default ranking; under an unpinned dependency one method returned its input column order on every case (\cref{sec:rq2-defect}).

(iii) Input tables should be validated before release; a duplicated header in 50 tables produced a CI-supported reversal that disappeared once removed.

(iv) Coverage and a full-case denominator should accompany every score; reported values that drop failed cases cannot be matched by a reader who keeps them (\cref{sec:rq2-denominator}).

(v) Per-case outputs of every reported method should be released, so that readers can compute paired differences and intervals, which none of the papers we could check allows (\cref{sec:bg-survey}).

The current upstream release of \rcaeval includes fixes for (ii) and (iii).

\subsection{Relation to Benchmark-Difficulty Critiques}
\label{sec:disc-fang}

\citet{fang2025simplerca} show that a simple rule-based method reaches the reported state of the art on existing RCA benchmarks and propose a harder benchmark.
We hold the benchmarks fixed and ask which of the comparisons drawn from them hold across their subsystems and which part of a score the released pipeline produced.
The two concerns are independent and their remedies combine: a harder benchmark still needs its comparisons scoped and their provenance stated, and both stated on an easy benchmark still overstate what methods can do.
Where the results overlap they agree: the authors' \simplerca code is also best on RE2-TT in our runs.

\section{Related Work}
\label{sec:related}

\textbf{Benchmark validity in machine learning.}
Pooled leaderboards have been criticized for hiding condition-specific behavior and for depending on which benchmarks are included~\citep{dehghani2021benchmark,bowman2021will,liang2022helm}.
Test-set replication and label errors can reorder methods~\citep{recht2019imagenet,northcutt2021pervasive}, and hidden stratification and underspecification describe models whose average performance hides failures on subgroups~\citep{oakdenrayner2020hidden,damour2022underspecification,sagawa2020distributionally}, and checklists for benchmark quality ask for documentation and reproducible releases~\citep{betterbench,pineau2021improving}.
We apply this view to RCA benchmarks, where subsystems are identifiable applications, per-case scores are available, and the release itself can be under-specified.

\textbf{Empirical software engineering.}
Guidelines for empirical research in software engineering ask that a study state its population, treatments, and analysis in enough detail to be repeated~\citep{kitchenham2002guidelines}, and catalogues of recurring problems in software analytics papers include inadequate reporting of data preparation and conclusions whose stability is never tested~\citep{menzies2019badsmells}.
In defect prediction, the research group explained more of the variation in reported performance than the classifier did~\citep{shepperd2014researcher}; cross-project experiments found that whether a predictor transfers depends on the projects involved~\citep{zimmermann2009crossproject}, a benchmark of cross-project approaches found their ranking to depend on the datasets and metrics used~\citep{herbold2018comparative}, and the model-validation technique was shown to change performance estimates and rankings~\citep{tantithamthavorn2017validation}.
We ask the same questions of RCA benchmarks: whether a comparison holds on each evaluated system, and which part of a score the evaluation procedure produced.

\textbf{RCA benchmarks and methods.}
Offline RCA benchmarks include \rcaeval~\citep{pham2024rcaeval}, \openrca~\citep{xu2025openrca}, \petshop~\citep{hardt2024petshop}, and LEMMA-RCA~\citep{zheng2024lemma}; AIOpsLab and ITBench evaluate agents against live systems, and Cloud-OpsBench against recorded state snapshots replayed through diagnostic tools~\citep{aiopslab,jha2025itbench,wang2026cloudopsbench}.
Audited methods span change-point detection (\baro~\citep{pham2024baro}), causal inference and discovery (\circa~\citep{li2022circa}, \rcd~\citep{ikram2022rcd}, \causalrca~\citep{causalrca}, \cfa~\citep{budhathoki2022causal}), anomaly scoring (\ediag~\citep{ediagnosis}), and trace analysis (\tracerca~\citep{li2021tracerca}, \microrank~\citep{yu2021microrank}).

\textbf{Evaluating RCA evaluation.}
\citet{pham2024howfar} compare causal-inference RCA methods on common data and find that no method stands out in all situations. We do not re-establish that winners differ; we ask, for the pairwise comparisons a released benchmark supports, on which subsystems a pooled lead holds and which part of it the evaluation pipeline produced, with matched per-case outputs that others can recheck.
\citet{fang2025simplerca} question benchmark difficulty (\cref{sec:disc-fang}).
Studies of LLM agents for cloud RCA analyze why agents fail~\citep{kim2026agentfail,riddell2026stalled}; they concern a different evaluation mode from ours.

\section{Threats to Validity}
\label{sec:threats}

\textbf{Construct validity.}
\rcaeval and \petshop score top-1 hits, while \openrca gives partial credit for component, reason, and time.
We compare differences within a family, except for the cross-family set (\cref{sec:rq1-cross}), where we compare only the signs of subsystem differences and do not average scores across families.
A reversal can be defined by the point estimate or by the CI; we report both, and the conclusions of \cref{sec:rq1} hold under either.

\textbf{Internal validity.}
Our results depend on running released code correctly, which is exactly where we found problems.
Our own first RE1 runs also read the raw exports; we noticed only because \baro's Avg@5 did not match its authors' numbers.
We guard against similar errors in four ways.
We anchor implementations to published numbers where possible (\baro on RE1, \simplerca on RE2 and RE3, and \tracerca and \microrank on RE2-TT when invalid outputs are dropped); we check every ranking against the input column order; we apply the fixes of the current upstream release; and we record the configuration of every run.
The \openrca wrapper that turns metric rankings into answers is our design; it is the same for every method, and queries where it falls back to a default answer score 0.
Because methods favour different kinds of metrics, the mapping from metric type to failure reason may suit some methods better than others, so our \openrca results hold under this wrapper and not necessarily under another.
The three probes have no parameter chosen on the data they are scored on: \maxz and \alertcount are fixed z-score rules, and \cdmin's thresholds were set on \openrca before this audit, so its \openrca results are not out of sample while its results on the other families are.
Two randomized methods are averaged over three seeds; the last row of \cref{tab:factors} gives what a single seed changes, and the per-seed values are in the replication package.

\textbf{External validity.}
We audit three benchmark families with $k=11$ subsystems in the main analysis; they are not a sample from a population of deployments, and we make no claim beyond them.
RE2 uses the same applications as RE1, so it tests carry-over between releases, not between systems; it differs from RE1 in collection, sampling, and recorded metric kinds as well as in faults, and the released data do not allow these to be separated.
Two \petshop scenarios have only 8 cases, and we report results without them.
Our conclusions are limited to offline benchmarks with matched cases; interactive agent benchmarks may behave differently.

\textbf{Conclusion validity.}
We examine many comparisons, and subsystem CIs are not adjusted for multiplicity; a few CI-supported reversals are expected by chance, which is why we report counts and do not interpret single reversals.
At a Bonferroni-adjusted level over the \NBonfComp subsystem comparisons between published methods in the main families, \NBonfCIAdjExcl of the \NBonfCIExcl subsystem CIs that exclude zero still do so, and \NBonfRevCI of the \NMainPubRevCI CI-supported reversals of RQ1 remains (\cfa vs \rcd on the high-traffic scenario).
The reversals of RQ2 have CIs excluding zero on both sides under both conditions.
The survey of \cref{sec:bg-survey} is a convenience sample of \SurveyN papers checked by one author on two dates through the links given in the papers; releases can change afterwards, and for \SurveyRelUndet papers a linked release could not be reached.

\section{Conclusion}
\label{sec:conclusion}

A pooled leaderboard reports one number per method and leaves unstated on which systems the advantage holds and how the score was produced.
In three RCA benchmark families, whether a pooled lead between published methods held on every subsystem was decided by a quantity the leaderboard withholds, the spread of the lead across subsystems: leads whose margin exceeded it held everywhere, most leads below it reversed, and every wrong choice on a held-out subsystem was of the second kind.
On a second release of the same applications, one method's response to newly recorded metric kinds accounted for five of the six comparisons that changed sign, and the input table, the handling of failed outputs, and the denominator of the released pipelines changed conclusions without raising an error.
Stating the subsystems and the spread next to every pooled comparison, naming the input table and the denominator, and releasing the per-case outputs behind it would let readers check what the leaderboard does not say.

\section*{Data Availability}
The replication package is available at \url{https://anonymous.4open.science/r/pooled-leaderboard-audit-replication-8DB8/}.
It contains the frozen per-case table with its SHA-256 digest, the run ledger, the run and analysis scripts, the adapted \simplerca code, the integrity-check diagnostics, the survey data, and additional result tables; a quick check regenerates every number, table, and figure of the paper from the per-case table in a few minutes, and a full rerun starts from the public benchmark data, which the package does not redistribute.

\bibliographystyle{ACM-Reference-Format}
\bibliography{references}

\end{document}